\documentclass{article}
\usepackage{stywhispers,amsmath,epsfig}
\usepackage{tocloft}
\usepackage{hyperref}
\usepackage{setspace}
\usepackage{float}
\usepackage{graphicx}

\usepackage{lastpage}
\usepackage{bbold}
\usepackage{placeins}
\usepackage{enumitem}
\usepackage{listings}
\usepackage{multirow} 
\usepackage{array}
\usepackage{lscape} 
\usepackage{longtable} 
\usepackage[T1]{fontenc}
\usepackage{fancyhdr,lipsum}
\usepackage{cite}
\usepackage{amsmath,amssymb,amsfonts}
\usepackage{algorithmic}
\usepackage{graphicx}
\usepackage{textcomp}
\usepackage{xcolor}

\usepackage{subcaption}

\usepackage{xurl}
\DeclareMathOperator{\Conv}{Conv}

\usepackage[export]{adjustbox}

\title{Homology features for X-Ray Hyperspectral Images Segmentation\thanks{Author Email: {\small fethi.harkat@univ-grenoble-alpes.fr}}}

\name{Fethi Harkat\textsuperscript{\large*} \textsuperscript{\large†},   Guillaume Gey\textsuperscript{\large*}, Val\'erie Perrier\textsuperscript{\large†}, Kévin Polisano\textsuperscript{\large†},}

\address{\textsuperscript{\large*}Detection Technology Plc, A Grid, Otakaari 5A, 02150 Espoo, Finland \\
\textsuperscript{\large†} Univ. Grenoble Alpes, CNRS, Grenoble INP, LJK, 38000 Grenoble, France}

\begin{document}

\maketitle

\begin{abstract}
Characterizing Multi-Energy X-ray Transmission (XRT) hyperspectral images presents significant challenges for traditional classification frameworks due to high-dimensional redundancy and non-linear spectral overlaps. Although manifold learning techniques such as UMAP enhance spectral representation by preserving non-linear correlations, they do not explicitly utilize the geometric structure of the embedded manifold for segmentation. This study introduces a hybrid architecture that integrates topological features derived from persistent homology with convolutional and graph neural networks. The approach computes Vietoris-Rips persistent homology on UMAP-projected superpixel point clouds to capture $H_1$ cyclic structures that encode spectral heterogeneity. These topological descriptors are combined with pixel-level CNN features through a broadcasting mechanism that propagates superpixel-level graph representations to full spatial resolution. Experimental results on XRT multi-energy imagery demonstrate that the proposed hybrid model significantly outperforms both CNN-only and superpixel-pooling fusion baselines, suggesting that topological features provide complementary discriminative information to local spectral-spatial descriptors.
\end{abstract}

\begin{keywords}
Hyperspectral imaging, X-ray detection, Topology, Manifold learning, Dimensionality reduction, Deep learning, Feature extraction, Computer Vision. 
\end{keywords}

\section{Introduction}
\label{sec:introduction}

Linear projection techniques such as Principal Component Analysis (PCA) \cite{jolliffe2016principal} do not capture the non-linear geometry of Hyperspectral Images (HSI) \cite{bachmann2005exploiting}, and our previous work showed that they also fail to represent the intrinsic geometry of XRT datasets \cite{harkat2025manifold}. The Uniform Manifold Approximation and Projection (UMAP) algorithm \cite{mcinnes2018umap, healy2024uniform} was shown to improve optical HSI classification over PCA \cite{lou2025hyperspectral}, and our previous work established that the manifold underlying XRT data exhibits non-linear correlations that can be characterized through topological approximation \cite{harkat2025manifold}. These results indicate that the topology of the data contains discriminative information that is not captured by standard spectral-spatial feature extractors.

Building upon this foundation, we propose a hybrid architecture that integrates tools from Topological Data Analysis (TDA), specifically persistent homology \cite{edelsbrunner2010computational, carlsson2009topology}, with Graph Neural Networks (GNNs) \cite{wu2020comprehensive, zhou2020graph} and Convolutional Neural Networks (CNNs). The high dimensionality and spectral redundancy of HSI data have motivated a shift from pixel-wise classification to region-based reasoning through graph representations \cite{liu2020cegcn}. In this framework, the image is abstracted as a graph where nodes correspond to image regions and edges encode spatial or spectral adjacency. Early approaches constructed graphs at the pixel level, treating each spectral vector as a node and connecting $k$-nearest neighbors in spectral space \cite{hong2020graph}. While preserving fine-grained spatial resolution, this approach suffers from computational intractability and susceptibility to spectral noise, as the graph connectivity becomes unreliable in high-dimensional, noisy spectral spaces \cite{liu2020cegcn}. Graph Convolutional Networks (GCNs) \cite{kipf2016semi} operate on this structure through message-passing mechanisms that propagate information across the graph, enabling the integration of local spectral features with global spatial context.

Computing persistent homology requires a sufficient number of data points to form meaningful topological structures. Operating at the pixel scale is impractical because individual pixels lack adequate spectral information to construct stable persistence diagrams. We therefore adopt a superpixel-based strategy. Following our previous methodology in  \cite{harkat2025manifold}, we first perform a UMAP projection on the image before grouping pixels into perceptually coherent superpixels using Simple Linear Iterative Clustering (SLIC)~\cite{achanta2012slic}. Each superpixel's spectral point cloud constitutes an independent metric space on which Vietoris-Rips persistent homology is computed via Ripser \cite{bauer2021ripser}. The resulting persistence statistics, combined with spatial descriptors, form node features for a GCN operating on the superpixel adjacency graph. A broadcasting mechanism propagates superpixel-level graph features back to individual pixels, where they are concatenated with pixel-level CNN features for segmentation.

Prior work has integrated topological summaries into machine learning pipelines through persistence images \cite{adams2017persistence}, landscapes \cite{bubenik2015statistical}, kernel embeddings \cite{reininghaus2015stable}, and direct graph architectures \cite{carriere2020perslay, horn2022topological}. However, the application of localized persistent homology to superpixel-based representations of hyperspectral data remains unexplored. Furthermore, existing hybrid CNN-GNN frameworks treat the two branches of the two neural networks as independent feature extractors whose outputs are combined only at the final layer, missing opportunities for intermediate information exchange \cite{hong2020graph, liu2020cegcn}. None of these approaches explicitly incorporate topological features into the GNN branch.

The key contributions of this paper are threefold. First, we demonstrate that persistent homology features computed on UMAP-projected superpixel point clouds provide complementary discriminative signal to standard spectral-spatial descriptors. Second, through systematic ablation, we identify a minimal subset of $H_1$ persistence statistics that optimizes the trade-off between feature dimensionality and segmentation performance. Third, we propose a broadcasting fusion strategy that preserves pixel-level spatial resolution and outperforms superpixel-pooling alternatives by a substantial margin. Experiments on XRT multi-energy imagery validate the proposed framework.

The remainder of this paper is organized as follows. Section~\ref{sec:methodology} details the proposed methodology, including UMAP projection, superpixel segmentation, topological feature extraction, and the hybrid CNN-GNN architecture. Section~\ref{sec:experiments} describes the experimental setup and datasets, then presents quantitative and qualitative results. Section~\ref{sec:conclusion} concludes with directions for future work.

\section{Methodology}
\label{sec:methodology}

\subsection{Overview}

\begin{figure}[ht]
    \centering
    \includegraphics[width=1\linewidth]{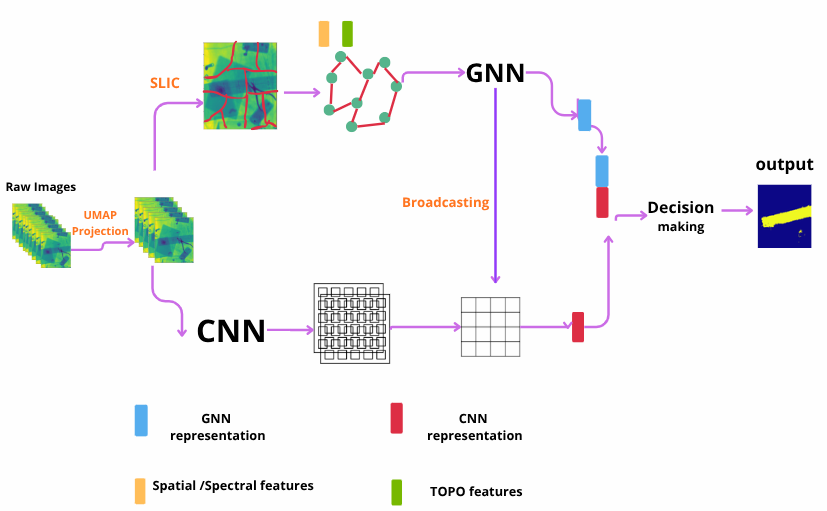}
    \caption{Pipeline of the proposed hybrid CNN+GNN method. The CNN branch (bottom) processes the UMAP-projected image at full pixel resolution, while the GNN branch (top) operates on superpixel-level topological and spatial features. A broadcasting mechanism fuses both representations before the final classification layer.}
    \label{fig:architecture}
\end{figure}

The proposed framework, illustrated in Fig.~\ref{fig:architecture}, consists of four stages. First, raw XRT hyperspectral cubes are projected onto a low-dimensional manifold using UMAP \cite{mcinnes2018umap}, fitted on a sample of training pixels, following the approach in \cite{harkat2025manifold}. Second, the projected image is segmented into superpixels. Third, topological and spatial features are extracted from each superpixel's spectral point cloud. Fourth, a dual-branch hybrid architecture fuses pixel-level CNN features with superpixel-level GNN features via a broadcasting mechanism, followed by a final classification layer. The first three stages are pre-processing steps; only the two branches and the classifier are trained. We detail below these different steps.

\subsection{Superpixel Segmentation and Graph Construction}

Raw ME-XRT images are acquired as three-dimensional cubes of shape $[H, W, C]$, where $H$ and $W$  denotes spatial dimensions, the detector pixel axis, and $C$ the energy band axis \cite{harkat2025manifold}. UMAP \cite{mcinnes2018umap}, fitted on a sample of training pixels, projects each pixel's $C$-dimensional spectral signature onto a $D$-dimensional manifold ($D \ll C$; here $C=16$ and $D=5$), yielding projected images of shape $[H, W, D]$ that preserve non-linear spectral correlations.

The UMAP-projected image is then segmented into $N$ perceptually coherent superpixels using SLIC \cite{achanta2012slic}, operating on the $D$ manifold channels. The resulting superpixels form the nodes of a graph $\mathcal{G} = (\mathcal{V}, \mathcal{E})$, where edges connect spatially adjacent regions. Thus, the graph comprises $N$ nodes instead of the original $H \times W$ pixels, reducing the spatial complexity from $\mathcal{O}(HW)$ to $\mathcal{O}(N)$ while preserving spatial coherence. The spectral dimension is embedded in the $D$-dimensional UMAP coordinates and subsequently encoded in the topological descriptors extracted per superpixel.
\subsection{Node Feature Extraction}
\label{sec:node-features}
For each superpixel $s_i$, let $\mathcal{P}_i \subset \mathbb{R}^D$ denote its UMAP-projected spectral point cloud. Vietoris--Rips persistent homology is computed via Ripser~\cite{bauer2021ripser}, yielding persistence diagrams $\mathcal{D}_0(\mathcal{P}_i)$ and $\mathcal{D}_1(\mathcal{P}_i)$. Diagram $\mathcal{D}_0$ captures connected components (0-dimensional features), while $\mathcal{D}_1$ captures 1-dimensional loops. From these diagrams we extract a pool of 30 candidate topological descriptors.

\subparagraph*{Topological descriptors (30 candidates).}
\label{topo}
For each birth--death pair $(b_j, d_j)$ in a persistence diagram, the persistence is $\ell_j = d_j - b_j$. Let $n_k = |\mathcal{D}_k(\mathcal{P}_i)|$ ($k=0,1$) denote the number of points in $\mathcal{D}_k$. The $H_1$-statistics are defined as:
{\small
\begin{align*}
\text{MaxPers} &= \max_j \ell_j, & \text{MinPers} &= \min_j \ell_j, \\
\text{MeanPers} &= \tfrac{1}{n_1}\sum_j \ell_j, & \text{StdPers} &= \sqrt{\tfrac{1}{n_1}\sum_j (\ell_j - \bar{\ell})^2}, \\
\text{SumPers} &= \sum_j \ell_j, & \text{MeanBirth} &= \tfrac{1}{n_1}\sum_j b_j, \\
\text{MeanDeath} &= \tfrac{1}{n_1}\sum_j d_j, & \text{MaxBirth} &= \max_j b_j, \\
\text{MaxDeath} &= \max_j d_j, & \text{Entropy} &= -\sum_j p_j \log p_j,
\end{align*}
}%
where $p_j = \ell_j / \sum_k \ell_k$, and $\text{PointsCount} = n_1$ is the 11th feature. The same 11 statistics are computed on $\mathcal{D}_0$ (denoted $H_0$\_*). Additionally, Betti numbers $\beta_0$ and $\beta_1$ are recorded at three fixed filtration scales $\epsilon_1 < \epsilon_2 < \epsilon_3$ (6 features), and two composite scores ($\text{Total\_Complexity}$, $\text{Topology\_Score}$) combine multi-scale topological indicators, yielding 30 candidates in total.

\subparagraph*{Spectral descriptors (25 dimensions).}
For each UMAP channel $d \in \{1,\dots,D\}$ (here $D=5$), the mean, standard deviation, and quartiles (Q1, Q2, Q3) of pixel intensities within the superpixel are computed, giving $5 \times 5 = 25$ dimensions.

\subparagraph*{Geometric descriptors (5 dimensions).}
A superpixel $s_i$ is a connected set of pixels; let $\mathcal{R}_i = \{(h,w) \in [H] \times [W] \mid \text{pixel } (h,w) \text{ belongs to } s_i\}$ denote the collection of pixel coordinates belonging to superpixel $s_i$. From this set we extract area $|\mathcal{R}_i|$ (the pixel count), perimeter, and:
\begin{align*}
\text{Eccentricity} &= \sqrt{1 - \frac{\lambda_{\min}}{\lambda_{\max}}}, \\
\text{Solidity} &= \frac{|\mathcal{R}_i|}{|\Conv(\mathcal{R}_i)|}, \\
\text{Orientation} &= \frac{1}{2} \arctan\left(\frac{2\Sigma_{12}}{\Sigma_{11} - \Sigma_{22}}\right),
\end{align*}
where $\lambda_{\max}, \lambda_{\min}$ are the eigenvalues of the coordinate covariance matrix $\mathbf{\Sigma}$ of the points in $\mathcal{R}_i$, and $\Conv(\mathcal{R}_i)$ denotes the convex hull of $\mathcal{R}_i$ (the smallest convex polygon containing all pixels of the superpixel). Thus $|\Conv(\mathcal{R}_i)|$ is the area of that convex hull, and the ratio $|\mathcal{R}_i|/|\Conv(\mathcal{R}_i)|$ measures boundary regularity ($1$ for a convex shape, $<1$ for irregular boundaries).

\subparagraph*{Feature vector.}
Each superpixel node is initially described by a 60-dimensional vector (30 topological + 25 spectral + 5 geometric). The procedure that selects the final 11 $H_1$ topological features retained for the GNN is detailed in the ablation study (Section~\ref{subsec:ablation}).

\subsection{Hybrid CNN-GNN Architecture}

\label{subsec:hybrid-arch}

As shown in Fig.~\ref{fig:architecture}, the architecture comprises two parallel branches whose outputs are fused before the final classification layer.

\paragraph*{CNN Branch.} The UMAP-projected image is processed by a fully convolutional network with no pooling, preserving spatial resolution. The architecture comprises three stages: a $3 \times 3$ convolution expanding the $D=5$ UMAP channels (where $D$ is the output dimension of the UMAP reduction from $C=16$ spectral bands) to 32 feature maps; a $3 \times 3$ convolution with a dilation factor of 2, achieving a $5 \times 5$ effective receptive field at 64 channels; and a $1 \times 1$ convolution projecting to 64 output channels. Each convolution is followed by batch normalization, and by a ReLU activation for the first two.
 
\paragraph*{GNN Branch.} The superpixel graph is constructed with 41-dimensional node features: 11 topological descriptors derived from $H_1$ persistence statistics (see Section~\ref{subsec:ablation}; maximum, mean, standard deviation, sum, and minimum persistence; mean and maximum birth and death scales; entropy; and point count) combined with 30 spatial descriptors (per-channel spectral statistics and geometric shape features). These node features are processed by a 3-layer GCN \cite{kipf2016semi} ($41 \to 64 \to 64 \to 2$), with batch normalization, ReLU activation and dropout ($p=0.3$) after the first two layers, producing a 2-dimensional output per superpixel that encodes the topological and structural signature of each region.
The source code is publicly available\footnote{\url{https://gricad-gitlab.univ-grenoble-alpes.fr/harkatf/homology-features-for-x-ray-hyperspectral-images-segmentation}}.

\subsection{Fusion}
The CNN branch outputs pixel-level features $\mathbf{F}_{\text{cnn}} \in \mathbb{R}^{H \times W \times 64}$. To combine these with the GNN's superpixel-level outputs $\mathbf{Z}^{(3)} \in \mathbb{R}^{N \times 2}$, where the superscript $(3)$ indicates the third (final) GCN layer and $N$ is the number of superpixels ($N \le 150$: SLIC is run with 150 target segments and superpixels of fewer than 5 pixels are discarded), we broadcast each superpixel's feature vector to all its constituent pixels. Let $\mathcal{S}: [H] \times [W] \to [N]$ map each pixel to its superpixel index. The broadcasted feature map $\mathbf{F}_{\text{broad}} \in \mathbb{R}^{H \times W \times 2}$ is defined by:
\begin{equation*}
\mathbf{F}_{\text{broad}, h,w,k} = \mathbf{Z}^{(3)}_{\mathcal{S}(h,w),k},
\end{equation*}
and set to zero for the pixels of discarded superpixels. The GNN output dimension is set to 2, the number of classes (background and target) of the binary segmentation task; these two channels are latent features learned jointly with the final classifier. The CNN and broadcasted GNN features are concatenated channel-wise and classified by a final $1 \times 1$ convolution:
\begin{equation*}
\widehat{\mathbf{Y}} = \text{Conv}_{66 \to 2}^{1 \times 1}\left( \text{Concat}\left[\mathbf{F}_{\text{cnn}}, \mathbf{F}_{\text{broad}}\right] \right) \in \mathbb{R}^{H \times W \times 2},
\end{equation*}
where $\widehat{\mathbf{Y}}$ contains the per-pixel logits of the two classes. The two branches and the final classifier are trained jointly, with gradients flowing through both branches simultaneously, whereas UMAP, SLIC and persistent homology are fixed pre-processing steps (see Fig.~\ref{fig:architecture}).
\paragraph*{Training}
We employ weighted cross-entropy at the pixel level to handle class imbalance:
\begin{equation}
\mathcal{L} = -\frac{\sum_{h,w} \omega_{Y_{h,w}} \log \sigma(\widehat{\mathbf{Y}}_{h,w})_{Y_{h,w}}}{\sum_{h,w} \omega_{Y_{h,w}}},
\end{equation}
where $\sigma$ is the softmax over the two classes, $Y_{h,w} \in \{0,1\}$ is the ground-truth label of pixel $(h,w)$, and the class weights $\omega_0 = 1.0$ and $\omega_1 = 5.0$ penalize false negatives on the target class.

\section{Experimental Setup and Results}
\label{sec:experiments}

\subsection{Dataset and Metrics}

We evaluate on the cigarette dataset from our prior work \cite{harkat2025manifold}, consisting of ME-XRT image patches depicting cigarette packs placed over luggage backgrounds, with binary ground-truth masks distinguishing target (cigarette) from background. This dataset is particularly suited for evaluating topological descriptors in XRT imagery because the partial superposition of two distinct materials creates spectral mixing regions at their interfaces, where $H_1$ cyclic structures in the UMAP-projected manifold encode material boundaries. Evaluation uses Accuracy and F1-score \cite{powers2011evaluation} at both the pixel and the superpixel level (a superpixel is positive if $\geq$70\% of its pixels are target).

\subsection{Feature Selection and Ablation}
\label{subsec:ablation}

The 11 topological descriptors retained in Section~\ref{sec:node-features} were selected by ablating the GNN branch \emph{in isolation}: the CNN branch was disabled during this phase, so the reported scores reflect pure graph-based classification without pixel-level CNN features. We kept the 30 spatial features (25 spectral statistics + 5 geometric descriptors) fixed and progressively added candidate topological features from the pool of 30 defined in Section~\ref{sec:node-features}, tracking the F1-score \cite{powers2011evaluation} on the test set over 10 runs per configuration (the data set is split beforehand into a training and a test set, same as in \cite{harkat2025manifold}).

Table~\ref{tab:ablation-gnn} reports the results. Topological features alone (30 candidates on raw HSI) yield the lowest performance (F1 = $0.283$), because persistent homology is invariant to pixel permutations within a superpixel and thus discards absolute spectral intensities and spatial coordinates. On raw HSI, adding the 11 selected $H_1$ statistics to the spatial features raises F1 from $0.685$ to $0.764$, whereas expanding to 19 or 30 topological features degrades it to $0.752$ and $0.725$, respectively: the additional candidates ($H_0$ statistics, Betti numbers at fixed scales \cite{edelsbrunner2010computational}, composite scores) increase node dimensionality from 41 to 60 without yielding extra discriminative power, diluting class-specific signals during graph convolution. The same 11 features bring a similar gain on UMAP-projected data, where F1 rises from $0.778$ to $0.823$, confirming a significant complementary signal.

\begin{table}
\centering
\caption{GNN ablation on raw HSI (16 bands) and UMAP-projected data (5 components). F1 (Target) denotes the F1-score on the positive (target) class; OA denotes Overall Accuracy. All scores are obtained with the GNN branch alone (no CNN fusion).}
\label{tab:ablation-gnn}
\small
\setlength{\tabcolsep}{3pt}
\begin{tabular}{|l|l|c|c|}
\hline
\textbf{Input} & \textbf{Node Features} & \textbf{F1 (Target)} & \textbf{OA} \\
\hline
Raw HSI & Topo only (30) & $0.283 \pm 0.011$ & $0.391 \pm 0.003$ \\
Raw HSI & Spatial only & $0.685 \pm 0.005$ & $0.912 \pm 0.006$ \\
Raw HSI & Spatial + 3 Topo & $0.683 \pm 0.008$ & $0.947 \pm 0.006$ \\
Raw HSI & \textbf{Spatial + 11 Topo} & $\mathbf{0.764 \pm 0.007}$ & $0.934 \pm 0.010$ \\
Raw HSI & Spatial + 19 Topo & $0.752 \pm 0.003$ & $0.948 \pm 0.002$ \\
Raw HSI & Spatial + 30 Topo & $0.725 \pm 0.005$ & $0.933 \pm 0.003$ \\
\hline
UMAP & Spatial only & $0.778 \pm 0.011$ & $0.935 \pm 0.009$ \\
\textbf{UMAP} & \textbf{Spatial + 11 Topo} & $\mathbf{0.823 \pm 0.003}$ & $\mathbf{0.956 \pm 0.006}$ \\
\hline
\end{tabular}
\end{table}

This pattern reflects the distinct nature of the descriptors. $H_0$ statistics correlate with superpixel size and density rather than material identity, whereas the selected $H_1$ descriptors quantify non-trivial cyclic structures in the spectral point cloud arising from spectral mixing in heterogeneous regions. The compact 11-feature subset thus captures the essential topological structure without redundant dimensionality.  \textbf{With and without UMAP.} Table~\ref{tab:ablation-gnn} also compares persistent homology computed directly in the original 16-band spectral space (raw HSI) and in the 5-dimensional UMAP space. The 11 $H_1$ features improve the spatial-only baseline in both spaces ($+0.079$ in F1 on raw HSI, $+0.045$ on UMAP), which shows that the cyclic structures they capture are already present in the original spectra and are not created by the embedding. UMAP nevertheless gives the best results, with F1 rising from $0.764$ to $0.823$ for the same 11-feature configuration: by unfolding the spectral manifold into a lower-dimensional metric space, it removes redundant and noisy dimensions and yields more stable persistence diagrams.

\subsection{Hybrid Architecture: Quantitative Results}
\label{subsec:hybrid-results}

\begin{table}
\centering
\caption{Hybrid CNN+GNN: fusion strategies vs. CNN baselines.}
\label{tab:hybrid-phase2}
\footnotesize 
\setlength{\tabcolsep}{3pt} 
\begin{tabular}{|l|c|c|c|c|}
\hline
& \multicolumn{2}{c|}{\textbf{Superpixel}} & \multicolumn{2}{c|}{\textbf{Pixel}} \\
\hline
\textbf{Method} & \textbf{Acc.} & \textbf{F1} & \textbf{Acc.} & \textbf{F1} \\
\hline
CNN (Pixel-wise) & $0.9605$ & $0.8467$ & $0.9433$ & $0.8245$ \\
& $\pm 0.003$ & $\pm 0.007$ & $\pm 0.004$ & $\pm 0.008$ \\
\hline
CNN (Superpixel-wise) & $0.9387$ & $0.8201$ & $0.9317$ & $0.8144$ \\
& $\pm 0.005$ & $\pm 0.009$ & $\pm 0.006$ & $\pm 0.009$ \\
\hline
Hybrid (Pooling) & $0.9589$ & $0.8720$ & $0.9496$ & $0.8569$ \\
& $\pm 0.003$ & $\pm 0.006$ & $\pm 0.004$ & $\pm 0.007$ \\
\hline
\textbf{Hybrid (Broadcasting)} & $\mathbf{0.9802}$ & $\mathbf{0.9329}$ & $\mathbf{0.9640}$ & $\mathbf{0.8945}$ \\
& $\mathbf{\pm 0.002}$ & $\mathbf{\pm 0.003}$ & $\mathbf{\pm 0.003}$ & $\mathbf{\pm 0.005}$ \\
\hline
\end{tabular}
\end{table}

Table~\ref{tab:hybrid-phase2} compares the two fusion strategies against CNN-only baselines. The broadcasting hybrid achieves the highest scores across all metrics, surpassing the best CNN baseline by $+0.086$ in superpixel F1 and $+0.070$ in pixel F1. Even the pooling hybrid outperforms both CNN variants, though it trails broadcasting by $-0.061$ in superpixel F1.

These gains show that persistent homology features carry useful information for XRT segmentation. The CNN captures local spectral-spatial patterns in the UMAP embedding, but it does not characterize the global structure of the spectral manifold. The GNN supplies this through $H_1$ statistics that quantify cyclic structures in each superpixel's spectral point cloud. These cycles encode spectral mixing, the patterns that appear when the X-ray beam crosses multiple materials, and are not visible to convolutional filters alone. The hybrid model therefore outperforms both CNN baselines because topological descriptors capture patterns that local features miss.

The gap between broadcasting and pooling comes from spatial resolution. Broadcasting sends each superpixel's topological descriptor to all its pixels, letting the final classifier mix local CNN features with global topological information at full resolution. Under pooling, the CNN map is collapsed to superpixel averages before fusion, destroying fine spatial detail. Broadcasting is also more robust to segmentation errors: a bad superpixel poisons all its pixels under pooling, but under broadcasting the wrong topological vector only affects 2 of the 66 channels, which the CNN can partially correct.

These results extend our previous work~\cite{harkat2025manifold}, which relied on the UMAP projection alone: adding $H_1$ statistics through the GNN branch yields clear further gains, confirming that the topological structure of the spectral manifold carries discriminative information beyond the UMAP coordinates themselves.

\begin{figure}
    \centering
    \includegraphics[width=1\linewidth]{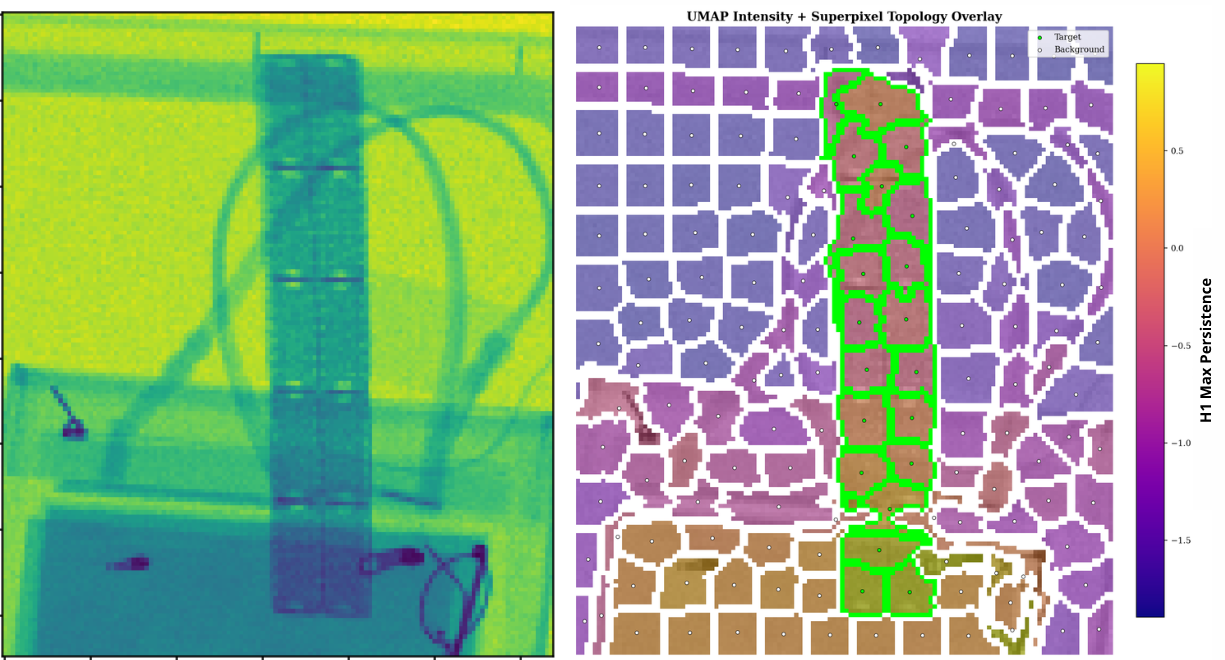}
    \caption{$H_1$ maximum persistence per superpixel overlaid on UMAP intensity. Warm colors indicate high persistence in heterogeneous regions; cool colors indicate homogeneous regions.}
    \label{fig:topology_boundary}
\end{figure}

Figure~\ref{fig:topology_boundary} supports this: superpixels with high $H_1$ maximum persistence (warm colors) correspond to regions where the spectral point cloud has complex structure, while homogeneous regions show low persistence (cool colors). The topological signal is strongest where spectral mixing is most pronounced.

\section{Conclusion}
\label{sec:conclusion}
This study investigated the integration of persistent homology features into a hybrid CNN-GNN framework for multi-energy XRT hyperspectral image segmentation. Our findings demonstrate that 1-dimensional persistent homology statistics derived from UMAP-projected superpixel point clouds effectively encode the non-linear structure and spatial complexity of the spectral manifold.
In particular, these topological descriptors capture local spectral mixing at material interfaces, providing complementary discriminative information that standard spectral-spatial features miss. Future works may focus on a deeper theoretical and empirical exploration of topological descriptors in X-ray hyperspectral imaging, investigating why persistent homology is particularly suited to modeling material boundaries, extending the framework to complex multi-class scenarios, and exploring adaptive superpixel partitioning strategies.

\bibliographystyle{IEEEbib}
\bibliography{bib}

\end{document}